\pdfoutput=1
\documentclass[conference]{IEEEtran}
\usepackage[pdftex]{graphicx}
\usepackage{amsmath}
\usepackage[caption=false,font=footnotesize]{subfig}
\usepackage{url}
\usepackage{fancyhdr}
\usepackage{hyperref}
\fancypagestyle{firstpage}{
  \fancyhf{}
  
  \fancyfoot[L]{{\em This paper was published in the Proceedings of the 2018 IEEE 12th International 
Conference on Semantic Computing (ICSC), pages 148-155, Laguna Hills, CA, USA, January 2018. \copyright 2018 IEEE\\
Link to article abstract in IEEE Xplore: http://dx.doi.org/10.1109/ICSC.2018.00029}}
}
\begin{document}
%
\title{Support Vector Machine Active Learning Algorithms with Query-by-Committee versus Closest-to-Hyperplane Selection}

\author{\IEEEauthorblockN{Michael Bloodgood}
\IEEEauthorblockA{Department of Computer Science\\
The College of New Jersey\\
Ewing, NJ 08628 USA\\
mbloodgood@tcnj.edu}}


\pagenumbering{gobble}

\maketitle

\thispagestyle{firstpage}

\begin{abstract}
This paper investigates and evaluates support vector machine active learning algorithms for use with imbalanced datasets, which commonly arise in many applications such as information extraction applications.
Algorithms based on closest-to-hyperplane selection and query-by-committee selection are combined with methods 
for addressing imbalance such as positive amplification based on prevalence statistics from initial random samples. 
Three algorithms (ClosestPA, QBagPA, and QBoostPA) are presented and carefully evaluated on datasets for text classification and relation extraction. 
The ClosestPA algorithm is shown to consistently outperform the other two in a variety of ways and insights are provided as to why this is the case.
\end{abstract}

%
\IEEEpeerreviewmaketitle

\section{Introduction}

The use of active learning has received a lot of interest for reducing annotation costs for text and speech processing applications \cite{mairesse2010, bloodgood2010ACL, lee2012, miura2016, hantke2017, mishler2017ICSC}. 
Many applications have the following three characteristics:
\begin{enumerate}
\item they have imbalanced data sets,
\item training data annotation is a burden, and
\item support vector machines (SVMs) are able to train high-performing systems for the application.
\end{enumerate}
Two examples of such applications are Text Classification (TC) and Relation Extraction (RE).

Characteristics 2 and 3 suggest the use of AL-SVM (Active Learning (AL) with Support Vector Machines). 
Previous work has presented an AL-SVM algorithm that selects (i.e., requests labels for) the examples that are closest to the current model's hyperplane \cite{tong2002,tong2001,campbell2000,schohn2000}. 
This ``closest"-based algorithm has been shown to need modification for imbalanced data situations \cite{bloodgood2009a}.
Previous work has presented a method for adapting to imbalanced data situations in the context of AL-SVM by using asymmetric cost factors during model training \cite{bloodgood2009a}.
The asymmetric cost model has been shown to be most effective when the model is based on prevalence statistics from an unbiased initial sample of data 
and serves as positive amplification for the minority positive examples.\footnote{It is typical for many applications that positive target examples are the minority class and are overwhelmed in number by negative examples. Sometimes these types of settings are referred to colloquially as {\em needle-in-the-haystack} settings.}
This method of dealing with imbalance during AL-SVM is denoted InitPA.
In this paper we refer to the algorithm that combines closest-to-hyperplane selection with InitPA cost modeling as ClosestPA.

Query by Committee (QBC) is another active learning algorithm that has been shown to be effective in a number of settings \cite{seung1992,abe1998,melville2004,mccallum1998}. 
Practical ways to build the committees include using Boosting and Bagging \cite{abe1998}. 
For imbalanced data situations, these algorithms will also benefit from a suitably adapted version of the InitPA cost modeling technique. 
Let the adapted algorithms be called QBoostPA and QBagPA.

This paper carefully compares ClosestPA, QBoostPA, and QBagPA for alleviating the training data 
annotation burden for applications that have the three previously mentioned characteristics.  
Experimental results are provided for multiple applications and datasets with different levels of naturally occurring imbalance, as might be encountered in realistic data mining settings.
ClosestPA is shown to be superior to the other two choices and insights are provided as to why this is the case.

Section~\ref{background} explains the algorithms in more detail, section~\ref{evaluation} contains empirical evaluation of the algorithms, 
section~\ref{discussion} contains insights as to why ClosestPA outperforms the other methods, section~\ref{relatedWork} contains related work, and section~\ref{conclusions} concludes.

\section{Algorithms} \label{background}

This section presents the algorithms that are investigated. 

\subsection{Active Learning with SVMs and Positive Amplification}

SVMs \cite{vapnik1998,cristianini2004} are learning systems that learn linear functions for classification. A statement of the optimization problem solved by soft-margin SVMs that enables the use of asymmetric cost factors is the following:

Minimize: 
\begin{equation}
\frac{1}{2}\|\vec{w}\|^{2} + C_{+}\sum_{i:y_{i}=+1}\xi_{i} + C_{-}\sum_{j:y_{j}=-1}\xi_{j}
\vspace*{-.5cm}
\end{equation}

Subject to: 
\begin{equation}
\forall k : y_k\left[\vec{w} \cdot \vec{x}_k + b \right] \ge 1 - \xi_{k} \enspace ,
\vspace*{-.2cm}
\end{equation}
where $(\vec{w},b)$ represents the hyperplane that is learned, $\vec{x}_k$ is the feature vector for example $k$, 
$y_k \in \{-1,+1\}$ is the label for example k, $\xi_k = max(0,1-y_k(\vec{w_k} \cdot \vec{x}_k + b))$ is the slack
variable for example k, and $C_+$ and $C_-$ are user-defined cost factors. 

\begin{figure}
\centering
\includegraphics[width=2.5in]{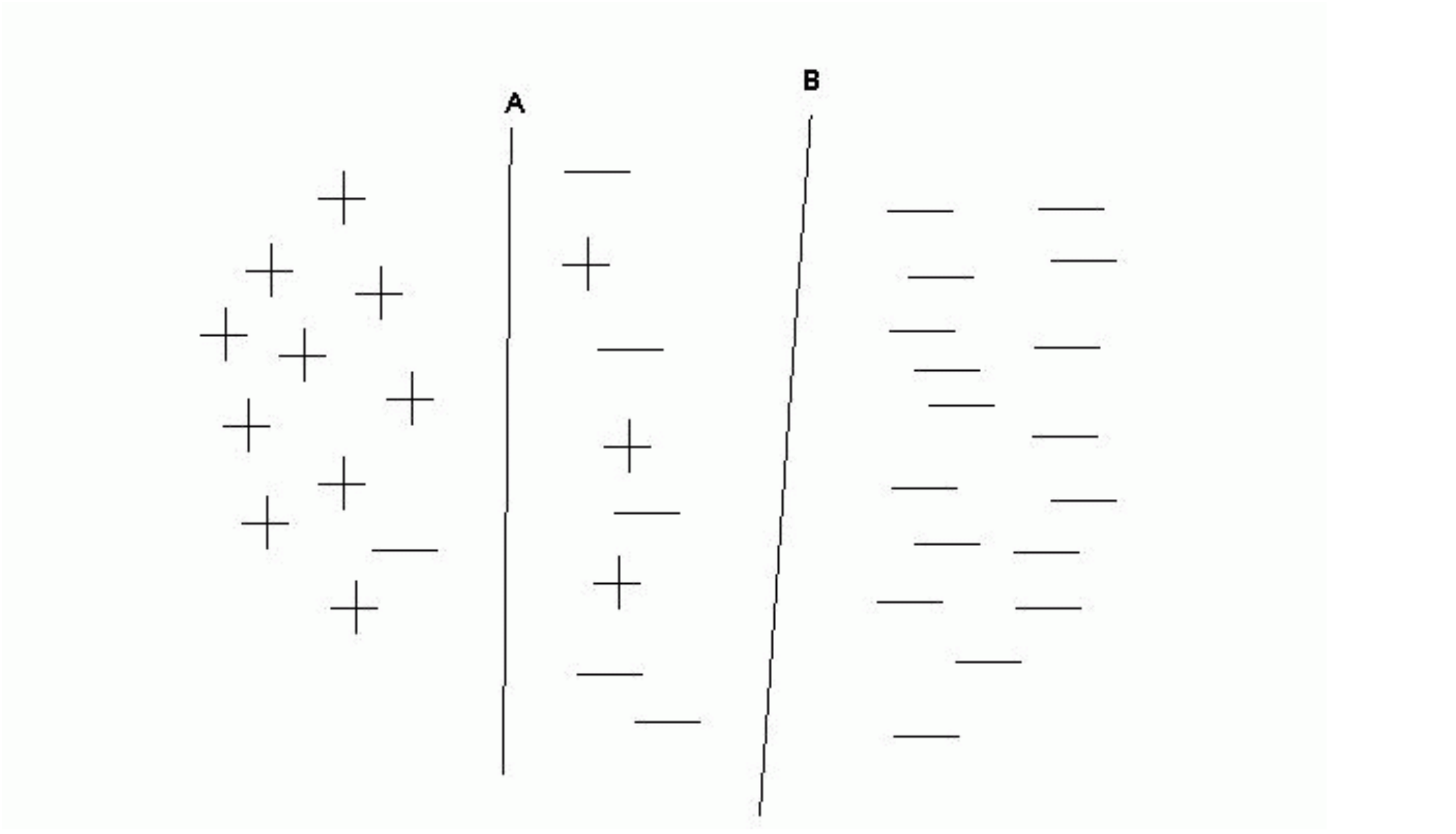}
\vspace{-.2cm}
\caption{Hyperplane B trained with a higher PA than hyperplane A trained with.}
\vspace{-.5cm}
\label{f:hyperplanes}
\end{figure}

$C_+$ and $C_-$ are user-defined cost factors that trade off separating the data with a large margin and misclassifying training examples. $C_+$ is the cost of misclassifying a positive training example and $C_-$ is the cost of misclassifying a negative training example.
Let PA=$\frac{C_+}{C_-}$. PA stands for Òpositive amplification.Ó PA is often allowed to default to 1 (i.e. $C_+ = C_-$), which we refer to as ÒNoPAÓ. For imbalanced datasets, it is prudent to actively set PA to be greater than one if positive examples are the minority class and less than one if positive examples are the majority class. Increasing PA typically shifts the learned hyperplane such that recall is increased and precision is decreased (see \figurename~\ref{f:hyperplanes} for a hypothetical example). A cost model that sets the PA well can have a significant beneficial impact on F measure\footnote{F measure is the standard metric used for evaluating performance of systems that perform search such as Text Classification and Relation Extraction systems. F measure is defined as the harmonic mean of precision and recall. Precision is the proportion of predicted positives that are indeed true positives. Recall is the proportion of positives that are predicted by the system as positive.}.

Morik et al. \cite{morik1999} describe how to set the PA effectively for passive learning from imbalanced datasets by using a cost model based on the distribution of negative and positive training examples. Bloodgood and Vijay-Shanker \cite{bloodgood2009a} explain that this distribution is skewed during AL-SVM and show how to integrate asymmetric cost factors into AL-SVM for imbalanced datasets by estimating PA from a small initial sample that is unbiased. 
Since we are interested in evaluating active learners for imbalanced datasets, we will use the technique of \cite{bloodgood2009a} to integrate asymmetric cost factors (i.e., PA $\ne 1$) into the active learners that we evaluate. This technique is referred to as the InitPA technique.

In order for the PA to be set appropriately, we want the proportion of positive instances in the small initial set of labeled
data to be an accurate estimate of the proportion of positive instances in the entire original pool of data. 
One can determine the sample size required to estimate the proportion of positives in a finite population to within sampling error $e$ with a desired level of confidence using the approach that is described in the next few paragraphs.

The approach is based on approximating a
Binomial distribution with a Normal distribution. A rule of thumb used by statisticians is that this 
approximation is justified only when the sample has at least 5 positive and at least 5 negative examples. 
In most situations, the sample sizes considered are substantial enough to 
meet the conditions for using the 
normal approximation. When the sampling distribution of the proportion can be assumed to be approximately normally 
distributed, we have
\begin{equation} \label{zInfinitePopulation}
z = \frac{p_S - p}{\sqrt{\frac{p(1-p)}{n}}}, 
\end{equation}
where $p_S$ = sample proportion,
$p =$ population proportion,
$n =$ sample size, and 
$z =$ z-score for standard normal distribution.
Multiplying both sides of Equation~\ref{zInfinitePopulation} by $\sqrt{\frac{p(1-p)}{n}}$, 
we obtain
\begin{equation} \label{zInf2}
z\sqrt{\frac{p(1-p)}{n}} = p_S - p.
\end{equation}
Define the sampling error $e$ as the difference between the sample estimate $p_S$ and the population parameter $p$. 
Then, substituting $e$ into Equation~\ref{zInf2}, we have
\begin{equation}
e = z \sqrt{\frac{p(1-p)}{n}},
\end{equation}
where $e = p_S - p$.
Solving for $n$, we obtain
\begin{equation} \label{nInfPopulation}
n = \frac{z^2 p (1-p)}{e^2}.
\end{equation} 

So to determine $n$, we need to know values for $z$, $e$, and $p$. 
The value for $z$ can be determined by looking up in a standard normal table the appropriate z-score corresponding to 
the level of confidence that is desired. 
The sampling error $e$ is the amount of error one is willing to accept in estimating the population proportion. 
The population parameter $p$ is what we are trying to estimate so we don't know its exact value. On pages 263-264 of 
\cite{berenson}, two possibilities are suggested. If past information or relevant experience is available that enables 
an educated estimate of $p$, then that estimate may be used. If past information or relevant experience is not available, 
then $p = 0.5$ can be used since $p=0.5$ will never underestimate $n$. This can be seen from Equation~\ref{nInfPopulation} where the product $p(1-p)$ occurs in the numerator and noting that the product $p(1-p)$ is maximized at $p=0.5$. 

When sampling from a population of finite size, as is the case for pool-based active learning, a finite population correction 
factor has to be used. Instead of Equation~\ref{zInfinitePopulation}, we have
\begin{equation} \label{zFinitePopulation}
z = \frac{p_S - p}{\sqrt{\frac{p(1-p)}{n}}\sqrt{\frac{N-n}{N-1}}},
\end{equation}
where $p_S$ = sample proportion,
$p =$ population proportion,
$n =$ sample size, 
$z =$ z-score for standard normal distribution, and
$N =$ size of the finite population (in our case the size of the original unlabeled pool of examples).
Multiplying both sides of Equation~\ref{zFinitePopulation} by 
$\sqrt{\frac{p(1-p)}{n}} \sqrt{\frac{N-n}{N-1}}$ and substituting $e$ for $p_S-p$, 
we obtain
\begin{equation}
z\sqrt{\frac{p(1-p)}{n}}\sqrt{\frac{N-n}{N-1}} = e.
\end{equation}

Solving for $n$, we have 
\begin{equation} \label{nFinitePopulation}
n = \frac{z^2 p (1-p)}{e^2} \frac{N-n}{N-1}.
\end{equation}

Note that the first term on the right-hand side of Equation~\ref{nFinitePopulation} is the estimate of the sample size that would be obtained using Equation~\ref{nInfPopulation}, 
which assumed an infinite population and thus did not employ a finite population correction factor. 
Define $n_0$ to be this uncorrected sample size estimate, i.e., 
\begin{equation} \label{nZero}
n_0 = \frac{z^2 p (1-p)}{e^2}.
\end{equation}
Substituting $n_0$ into Equation~\ref{nFinitePopulation}, we have
\begin{equation}
n = n_0 \frac{N-n}{N-1}.
\end{equation}

Solving for $n$ in terms of $n_0$ and $N$, we obtain 
\begin{equation} \label{correctedEstimate}
n = \frac{n_0 N}{N - 1 + n_0}.
\end{equation}

Then, summarizing, $n$ can be determined by using a two-step process. First determine the uncorrected sample size estimate, $n_0$, 
by using Equation~\ref{nZero}.
Then determine the finite population-corrected sample size estimate, $n$, by using Equation~\ref{correctedEstimate}.
In our case, $n$ can then be used to specify how many initial points should be labeled. 

As an example, carrying out the sample size determination 
computations on the AImed dataset shows that a size of 100 enables us to 
be 95\% confident that our proportion estimate is within 0.0739 of the true proportion. 
In our experiments, we used an initial labeled set of size 100.  

\subsection{Active Learning Selection Strategies}

A commonly used selection strategy used for AL-SVM is to use a closest-based example selection strategy \cite{tong2002,tong2001,campbell2000,schohn2000}. All of these previous works select the unlabeled examples that are closest to the current model's hyperplane and query for their labels. There are a few different theoretical motivations for using a closest-based selection strategy. An intuitive argument is that the examples that are closest to the hyperplane are the ones that the model is most unsure about and therefore, knowing the labels for those examples will provide the greatest benefit.

Another commonly used active learning selection strategy is Query by Committee (QBC) \cite{seung1992,abe1998,melville2004,mccallum1998}. This strategy works by using a committee of models and querying those unlabeled examples on which the committee disagrees the most about the label. A common way to form the committees is to use an ensemble learning method such as bagging or boosting \cite{abe1998}. When bagging is used, the resulting active learner is called QBag and when boosting is used, it's called QBoost.

We use the following notation to refer to particular AL algorithms:
\begin{itemize}
\item QBoostPA - QBoost with PA incorporated using InitPA technique,
\item QBagPA - QBag with PA incorporated using InitPA technique,
\item ClosestPA - closest-based strategy with PA incorporated using InitPA technique,
\item ClosestNoPA - a baseline method, this is the standard closest strategy where PA defaults to 1 (recall NoPA means PA=1),
\item RandomPA - a baseline method, this is random selection with PA set according to the cost model of \cite{morik1999}.
\end{itemize}

\section{Empirical Evaluation} \label{evaluation}

This section evaluates the algorithms described in section~\ref{background} on various datasets for relation extraction and text classification. These are applications where training data annotation has been shown to be a major bottleneck in the development of new systems and where successful active learning will make a significant contribution.

When these applications are modeled as binary classification tasks, they give rise to datasets that are imbalanced in the sense that there are more negative examples than positive examples. We experiment with datasets that exhibit a range of variations along the following dimensions: the total number of examples available, the level of class imbalance, the performance levels achieved with passive learning on all the data, and the {\em sparsity} \cite{campbell2000} of the solution.

On all of the datasets, SVMs have been shown to provide top-performing systems using a traditional passive learning setup. Thus, it is natural to explore AL-SVM for these datasets. 

The leading state of the art AL-SVM algorithms have never been evaluated head-to-head as they are in the following subsections. The results show that the ClosestPA active learner performs better than the other active learners in terms of data efficiency for achieving target F measure and in terms of F measure achieved at corresponding points along the learning curve.

Performance is not equally important at all points along the learning curve. At the beginning, when models are rapidly improving, performance is less important because you're not going to stop there. Also, near the end of an active learning simulation where all available data is eventually being labeled, the performance of different learners becomes similar. These points are not relevant because this is long after performance has leveled off and successful active learning would have stopped much earlier. In the middle, when performance is just starting to level off, is when effective stopping criteria will stop active learning and this is where performance differences matter most. ClosestPA is shown to significantly outperform the other algorithms at these important points along the learning curve.

When using active learning, it is important to strike a balance between using a small batch size to increase learning efficiency and a larger batch size to decrease runtime and increase annotator efficiency \cite{beatty2018ICSC}. Unless otherwise mentioned, our experiments in this paper use a batch size of 20.

\subsection{Evaluation Metrics} \label{evaluationMetrics}

Active learning is intended to reduce the amount of annotated data required to induce a top-performing model. One way to evaluate the effectiveness of active learners is to define the {\em target performance} as the performance that a baseline active learner can achieve on a given dataset, as determined by averaging its performance over the points on the learning curve corresponding to the last 100 training examples. Then we record the smallest number of labeled examples required by the baseline active learner to achieve the target performance. The data utilization ratio is the number of labeled examples required by an active learner to achieve the target performance divided by the number required by the baseline active learner. This metric reflects how efficiently the active learner is using the data and is similar to metrics used in previous research \cite{melville2004,abe1998}. 

F measure is the standard metric used to evaluate the performance of RE and TC systems. Therefore, we report data utilization ratio and other results based on F measure.

In addition to the data utilization metric, we also report graphs of the full learning curves during AL. In order to realize the potential performance enabled by the efficient selection of queries, an effective stopping criterion must be used \cite{schohn2000,bloodgood2013,bloodgood2009b}. Thus, we also report results based on performance measurements focused in the area of the learning curve where AL is expected to stop in practice. Finally, we use paired t tests at significance level 0.05 to perform statistical significance tests in the following subsections. Unless otherwise mentioned, the QBC approaches in the experiments use a committee size of five.

\subsection{Relation Extraction Experiments} \label{relationExtractionExperiments}

We conduct experiments on the AImed corpus, which was previously used for training protein interaction extraction systems in \cite{bunescu2005a,bunescu2005b,giuliano2006}. The protein interactions in AImed are examples of the domain-specific relations for which there now exist demand for RE systems to be built and for which manual annotation is expensive.

Consistent with previous work \cite{bunescu2005b,giuliano2006}, we cast RE as a binary classification task. Any pair of two annotated proteins occurring in the same sentence constitutes an instance. If the proteins are labeled as interacting, then it's a positive instance. Otherwise, it's a negative instance. All together, there are 5656 instances, of which 993 are positive and 4663 are negative. Note that the dataset is imbalanced, with positives constituting only 17.6\% of the instances.

The results reported below are based on the same 10-fold cross-validation setup and the same kernel as described in \cite{bloodgood2009a}. We use SVM$^{light}$ \cite{joachims1999} for our experiments.

Table~\ref{t:aimed} presents data utilization results for QBagPA,
QBoostPA, and ClosestPA when RandomPA is used as
the baseline active learner\footnote{The learners that use PA outperform those that don't use PA by large amounts.}. The numbers in the table show how many labeled points the various learners need to achieve the target F measure. Thus, lower numbers mean better data efficiency. Observe that QBagPA, QBoostPA, and ClosestPA all have better data utilization than RandomPA. Also note that ClosestPA has the best data utilization out of all of the active learners. The difference in data utilization between QBoostPA and QBagPA is not statistically significant, the difference between ClosestPA and QBoostPA is not statistically significant, and ClosestP A is statistically significantly better than QBagPA ($p < 0.05$).

\begin{table*}[t]
\renewcommand{\arraystretch}{1.3}
\caption{AIMed data efficiency with respect to RandomPA. Lower numbers indicate a more efficient method.}
\label{t:aimed}
\centering
\begin{tabular}{|c|c|c|c|c|c|c|}
\hline
Fold & Total Size & RandomPA & QBagPA & QBoostPA & ClosestPA & Target F Measure \\ \hline
1  & 5020 & 3660(1.00) & 1440(0.39) & 1920(0.52) & 1100(0.30) & 56.23 \\ \hline
2 & 5260 & 2020(1.00) & 1500(0.74) & 940(0.47) & 660(0.33) & 55.98 \\ \hline
3 & 5020 & 2920(1.00) & 2320(0.79) & 1420(0.49) & 2920(1.00) & 58.43 \\ \hline
4 & 4820 & 2360(1.00) & 3420(1.45) & 2460(1.04) & 2640(1.12) & 52.88 \\ \hline
5 & 5100 & 4180(1.00) & 1460(0.35) & 1220(0.29) & 440(0.11) & 46.07 \\ \hline
6 & 4820 & 860(1.00) & 1920(2.23) & 3700(4.30) & 400(0.47) & 55.74 \\ \hline
7 & 5160 & 3240(1.00) & 1740(0.54) & 1360(0.42) & 1760(0.54) & 60.80 \\ \hline
8 & 5180 & 3400(1.00) & 2700(0.79) & 2060(0.61) & 2840(0.84) & 59.83 \\ \hline
9 & 5180 & 5000(1.00) & 2080(0.42) & 1600(0.32) & 900(0.18) & 67.90 \\ \hline
10 & 5020 & 3400(1.00) & 2040(0.60) & 2840(0.84) & 2420(0.71) & 55.59 \\ \hline
Average & 5058 & 3104(1.00) & 2062(0.83) & 1952(0.93) & 1608(0.56) & 56.95 \\ \hline
\end{tabular}
\vspace*{-.2cm}
\end{table*}

\begin{figure}
\centering
\includegraphics[width=2.5in]{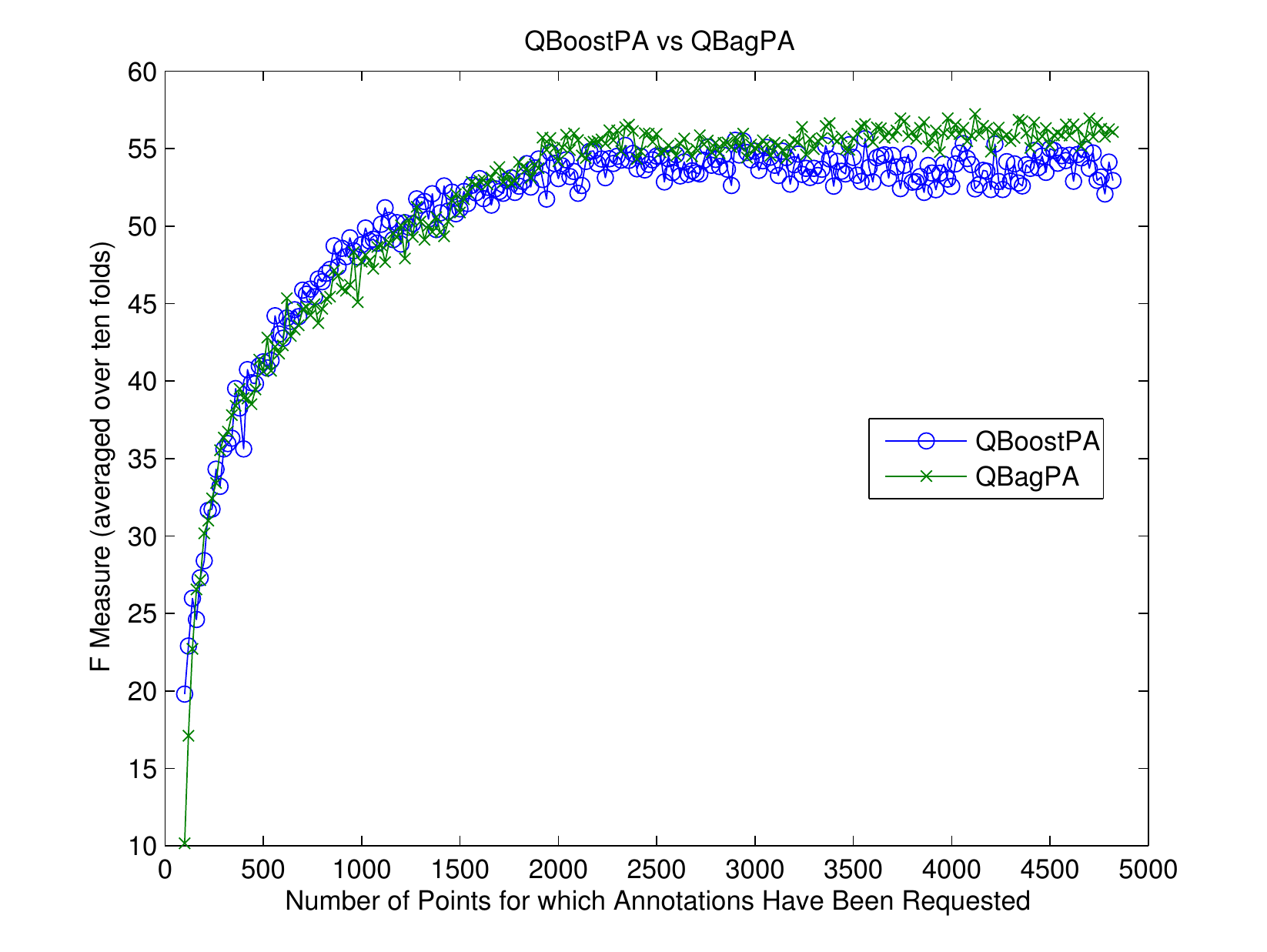}
\vspace{-.2cm}
\caption{QBoostPA vs QBagPA.}
\vspace{-.5cm}
\label{f:boostVsBag}
\end{figure}

\figurename~\ref{f:boostVsBag} plots the learning curves of QBoostPA and QBagPA. This plot and the rest of the plots in this subsection show performance averaged over the ten folds of cross validation. Observe that the difference in performance is small. In the very late stages, QBagPA performs a little better but this is largely irrelevant since the intention of AL is to stop querying for labels much earlier.

\begin{figure}
\centering
\includegraphics[width=2.5in]{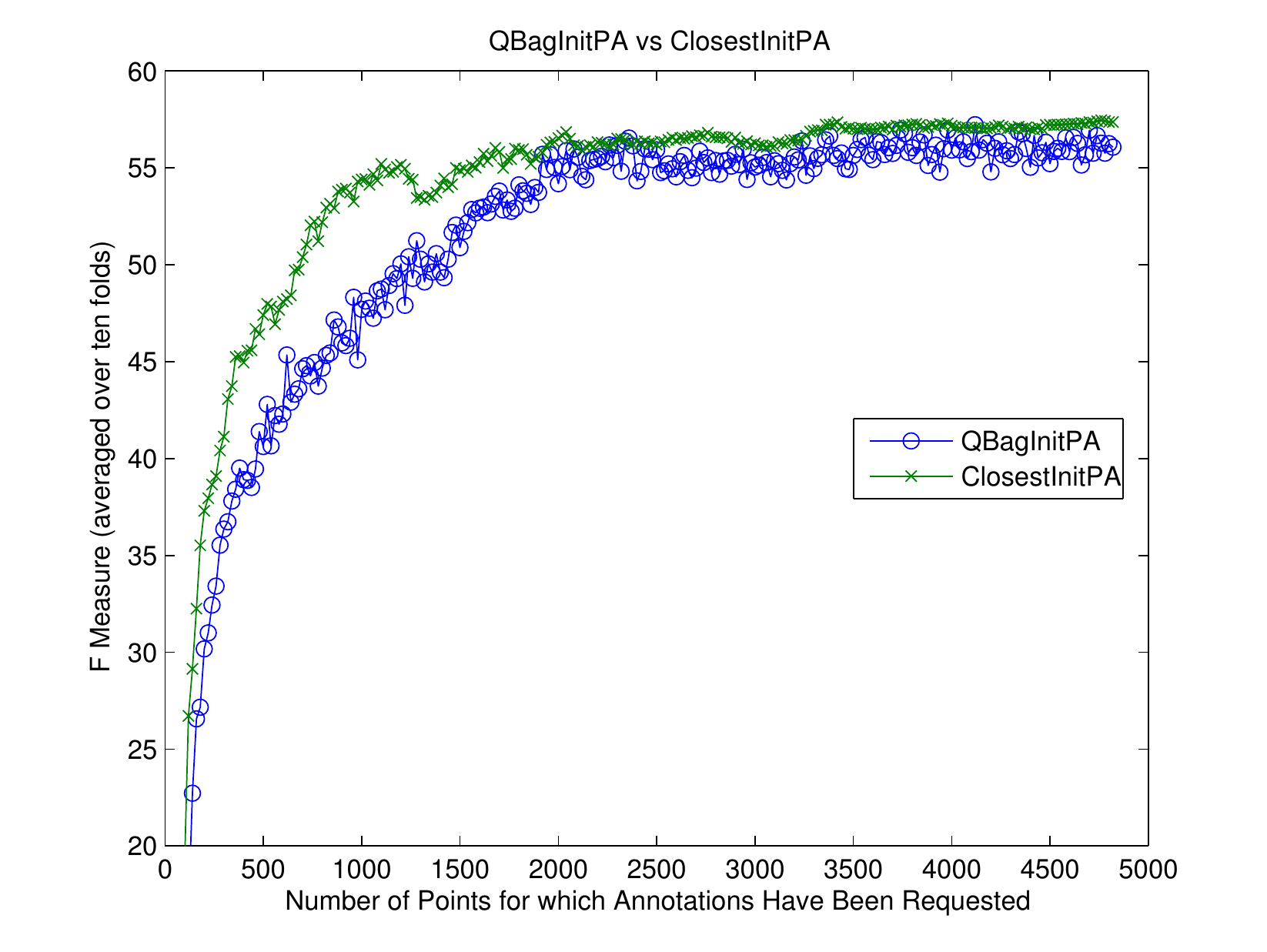}
\caption{QBagPA vs ClosestPA.}
\label{f:bagVsClosest}
\end{figure}

\begin{figure}
\centering
\includegraphics[width=2.5in]{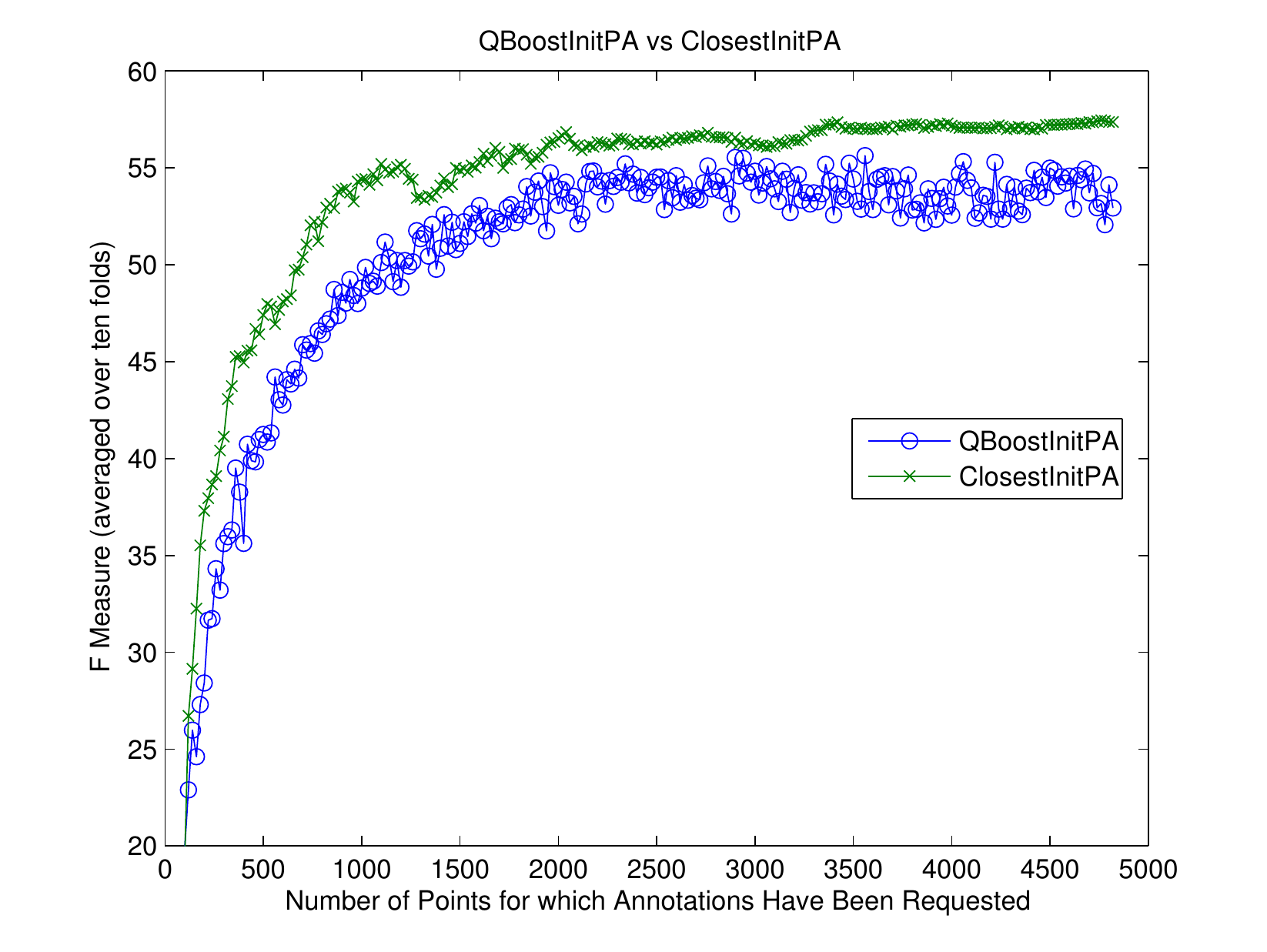}
\caption{QBoostPA vs ClosestPA.}
\label{f:boostVsClosest}
\end{figure}

\begin{figure}
\centering
\includegraphics[width=2.5in]{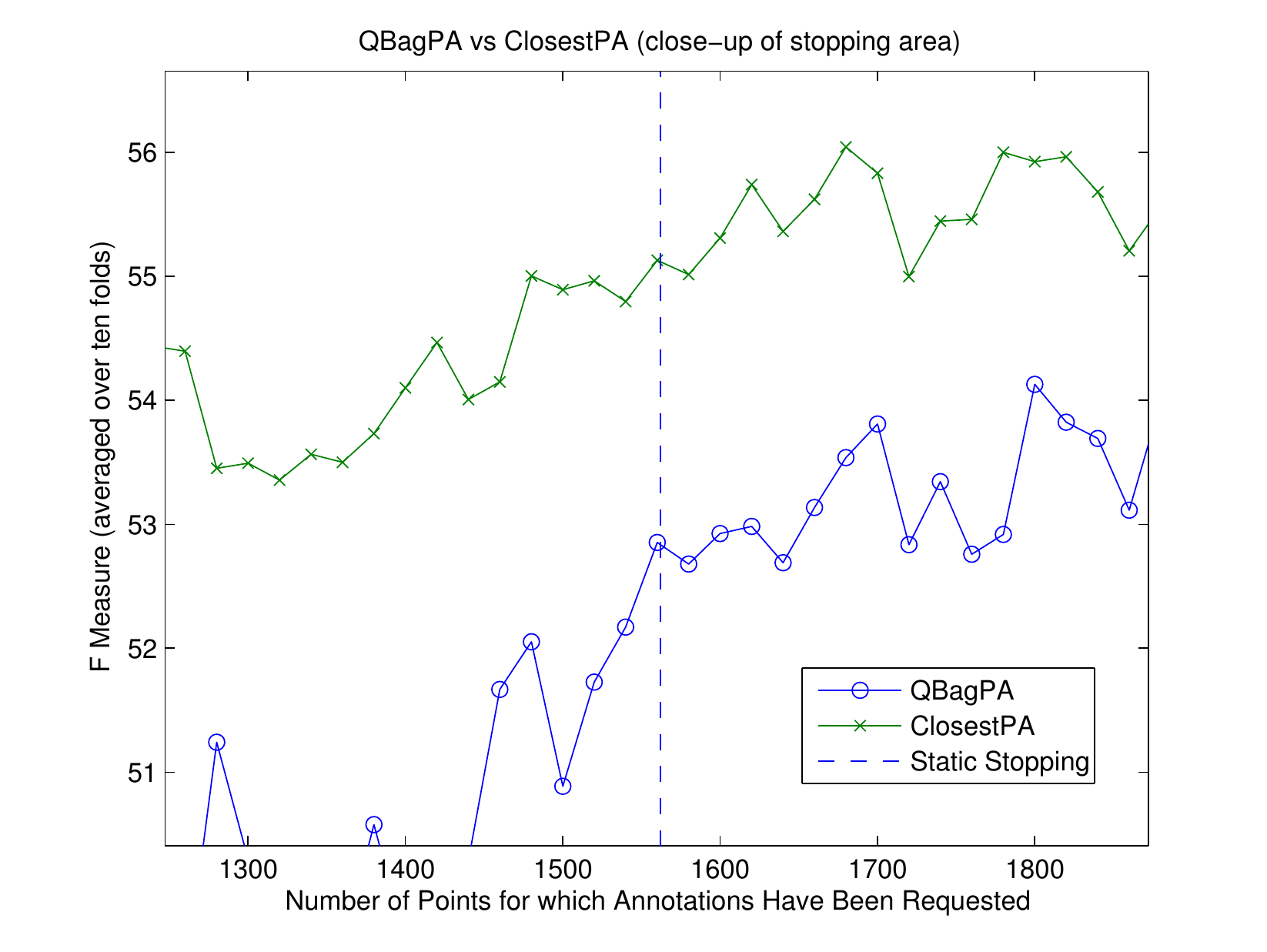}
\caption{QBagPA vs ClosestPA (close-up of stopping area).}
\label{f:bagVsClosestCloseup}
\end{figure}

\begin{figure}
\centering
\includegraphics[width=2.5in]{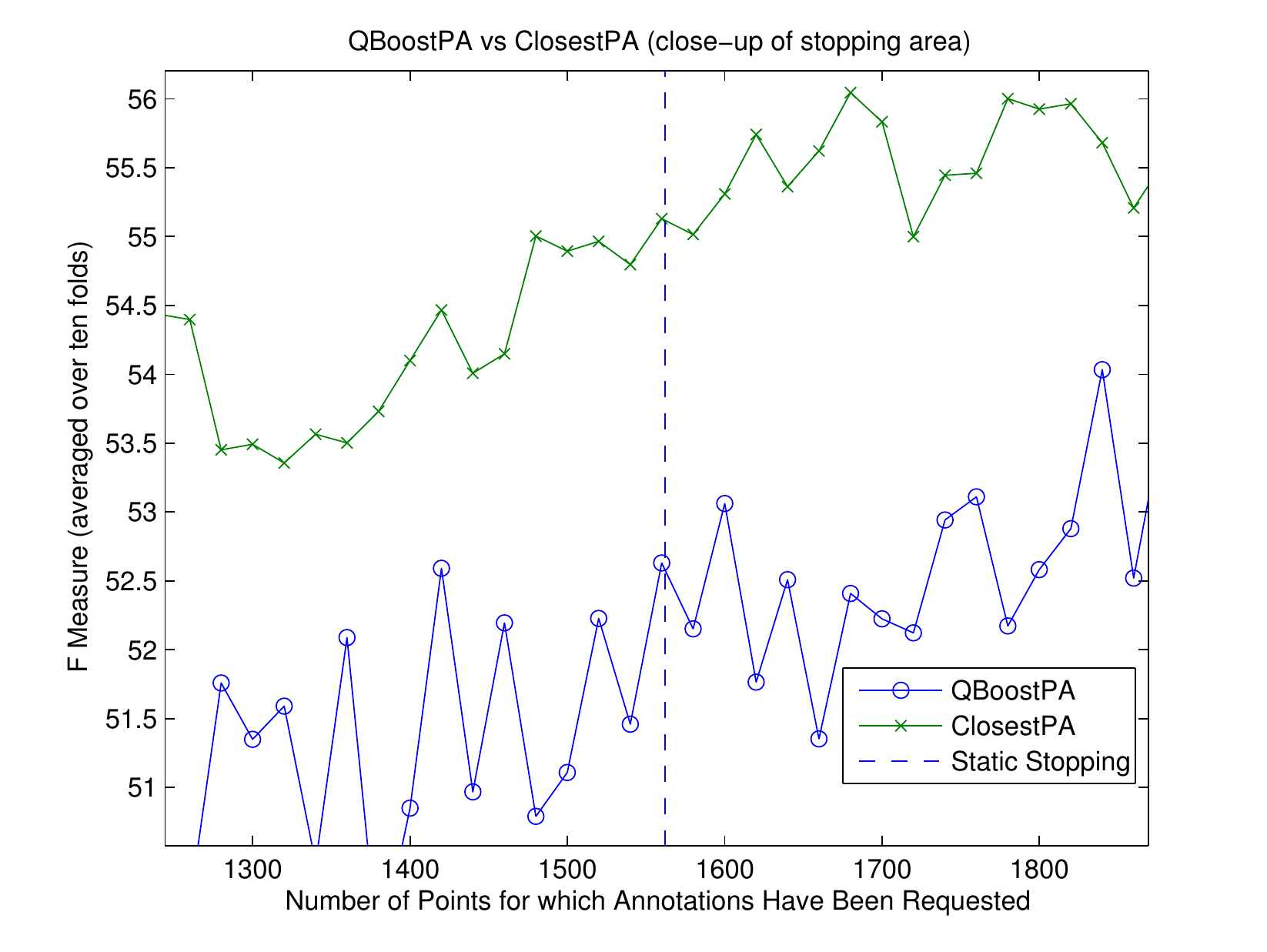}
\caption{QBoostPA vs ClosestPA (close-up of stopping area).}
\label{f:boostVsClosestCloseup}
\end{figure}

\figurename~\ref{f:bagVsClosest} and \figurename~\ref{f:boostVsClosest} plot the learning curves for ClosestPA versus QBagPA and for ClosestPA versus QBoostPA, respectively. In contrast to \figurename~\ref{f:boostVsBag}, note that in these figures the difference in performance is larger. In both cases, ClosestPA has stronger performance and especially in the most relevant area, the region where AL will stop in practice. \figurename~\ref{f:bagVsClosestCloseup} and \figurename~\ref{f:boostVsClosestCloseup} plot close-ups of the regions from \figurename~\ref{f:bagVsClosest} and \figurename~\ref{f:boostVsClosest} where it would be reasonable to stop querying for annotations. The dashed vertical line in these plots is where the stopping detection method from \cite{bloodgood2009b} indicates to stop active learning. Observe that ClosestPA outperforms both of the QBC-based approaches in this important region.

\subsection{Text Classification Experiments} \label{textClassificationExperiments}

All of our text classification datasets contain multiple categories. We treat them as binary classification tasks by using one-versus-the-rest classification for each category and averaging the results. 

The first TC dataset we use is the Reuters-21578 Distribution 1.0 ModApte
split\footnote{\url{http://www.daviddlewis.com/resources/testcollections/reuters21578}}. This    
dataset has 9603 training documents and 3299 test documents. Keeping with past practice
\cite{dumais1998,joachims1998}, we experiment with the 10 largest categories. We use a     
linear kernel with SVM$^{light}$ with binary features for each word that occurs in the       
training data at least three times.

This dataset has been used for previous AL-SVM research \cite{tong2002,schohn2000}. One of 
the main contributions of this past work was the ClosestNoPA algorithm.
Table~\ref{table:reutersDataEfficiency} reports on the performance of the active learners  
that use PA with respect to the
baseline ClosestNoPA active learner. Observe that ClosestPA again has the highest
performance out of all the active learners. Its performance is statistically significantly
better (p $<$ 0.05) than that of ClosestNoPA and QBoostPA.

Its performance is not statistically significantly better than QBagPA because
QBagPA has abysmal outlier performance on classification of documents in category `crude'.
One might think that this would help make ClosestPA be statistically significantly better
than QBagPA but it actually makes it harder because the sample estimate of the standard
deviation of the matched differences is now much higher. Removing the performance for the
`crude' category (where ClosestPA overwhelmingly outperformed QBagPA) and rerunning the
statistical significance test on the other nine categories indicates that ClosestPA is     
statistically significantly better than QBagPA. 

\begin{table*}[t]
\renewcommand{\arraystretch}{1.3}
\caption{Reuters data efficiency with respect to ClosestNoPA. Lower numbers indicate a more efficient method.}
\label{table:reutersDataEfficiency}
\centering
\begin{tabular}{|l|c|c|c|c|c|c|} \hline
Data Set& Total & ClosestNoPA & QBagPA & QBoostPA & ClosestPA & Target \\
& Size &&&&& F Measure \\ 
\hline
Earn & 9603 & 520(1.00) & 1080(2.08) & 1060(2.04) & 500(0.96) & 98.10 \\
\hline
Acq & 9603 & 1000(1.00) & 1340(1.34) & 1200(1.20) & 720(0.72) & 91.28 \\
\hline
MoneyFx & 9603 & 460(1.00) & 360(0.78) & 360(0.78) & 300(0.65) & 60.87 \\
\hline
Grain & 9603 & 540(1.00) & 620(1.15) & 620(1.15) & 420(0.78) & 85.18 \\
\hline
Crude & 9603 & 680(1.00) & 5280(7.76) & 300(0.44) & 280(0.41) & 80.07 \\
\hline
Trade & 9603 & 420(1.00) & 420(1.00) & 680(1.62) & 500(1.19) & 74.29 \\
\hline
Interest & 9603 & 440(1.00) & 380(0.86) & 500(1.14) & 260(0.59) & 61.76 \\
\hline
Ship & 9603 & 200(1.00) & 260(1.30) & 260(1.30) & 180(0.90) & 59.32 \\
\hline
Wheat & 9603 & 320(1.00) & 360(1.12) & 360(1.12) & 380(1.19) & 77.16 \\
\hline
Corn & 9603 & 1600(1.00) & 260(0.16) & 260(0.16) & 300(0.19) & 70.53 \\
\hline
Average & 9603 & 618(1.00) & 1036(1.76) & 560(1.10) & 384(0.76) & 75.86 \\
\hline
\end{tabular}
\end{table*}

In addition to the statistical significance, observe that ClosestPA requires practically   
significantly smaller numbers of labeled examples to obtain the same target F measure than 
the other algorithms require. 

The second text classification dataset we use is the Ohsumed collection \cite{hersh1994}.
This dataset has 6286 training documents and 7643 test documents. Keeping with past practice
\cite{joachims1997}, we report results for the five largest categories. We use a linear    
kernel with SVM$^{light}$ with binary features for each word that occurs in the training   
data at least three times. Data utilization ratio results on Ohsumed are reported in
Table~\ref{table:ohsumedDataEfficiency}. The results reinforce our findings on the other   
datasets, with ClosestPA outperforming the other algorithms again.

\begin{table*}[t]
\renewcommand{\arraystretch}{1.3}
\caption{Ohsumed data efficiency with respect to ClosestNoPA. Lower numbers indicate a more efficient method.}
\label{table:ohsumedDataEfficiency}
\centering
\begin{tabular}{|l|c|c|c|c|c|c|} \hline 
Data Set& Total & Closest & QBag & QBoost & Closest & Target \\
& Size & NoPA & PA & PA & PA & F Measure \\ 
\hline 
Immunologic& 6260 & 1780(1.00) & 860(0.48) & 720(0.40) & 400(0.22) & 44.20 \\
Diseases & & & & & &\\
\hline
Pathological & 6260 & 980(1.00) & 960(0.98) & 100(0.10) & 120(0.12) & 26.52 \\
Conditions, \ldots & & & & & &\\
\hline
Cardiovascular & 6260 & 860(1.00) & 1220(1.42) & 1440(1.67) & 1120(1.30) & 72.92 \\
Diseases & & & & & &\\
\hline
Neoplasms & 6260 & 580(1.00) & 1720(2.97) & 1780(3.07) & 520(0.90) & 70.08 \\
\hline
Digestive System& 6260 & 480(1.00) & 500(1.04) & 460(0.96) & 440(0.92) & 40.39 \\
Diseases& & & & & &\\
\hline
Average & 6260 & 936(1.00) & 1052(1.38) & 900(1.24) & 520(0.69) & 50.82 \\
\hline
\end{tabular}
\end{table*}

\subsection{Committee Size Experiments} \label{committeeSize}

The following experiments explore the potential for increasing the performance of the QBC methods by
increasing the committee size. The results show that the increases in performance are not worth the
extra computational expenses.  

A drawback of QBC-based approaches is that they are slower by a factor equal to the size   
of the committee. For AL applications, speed is important because we don't want annotators 
to have to wait while the system is deciding which examples to ask to be annotated. Thus,  
large committee sizes are impractical. With query by boosting, there is no way to          
parallelize the training of the different committee members. With query by bagging,        
parallelization is possible if resources are available. 

\begin{figure*}[t]
\centering
\subfloat[QBagPA with different committee sizes for category Immunologic Diseases from Ohsumed]{\includegraphics[width=5.7cm]{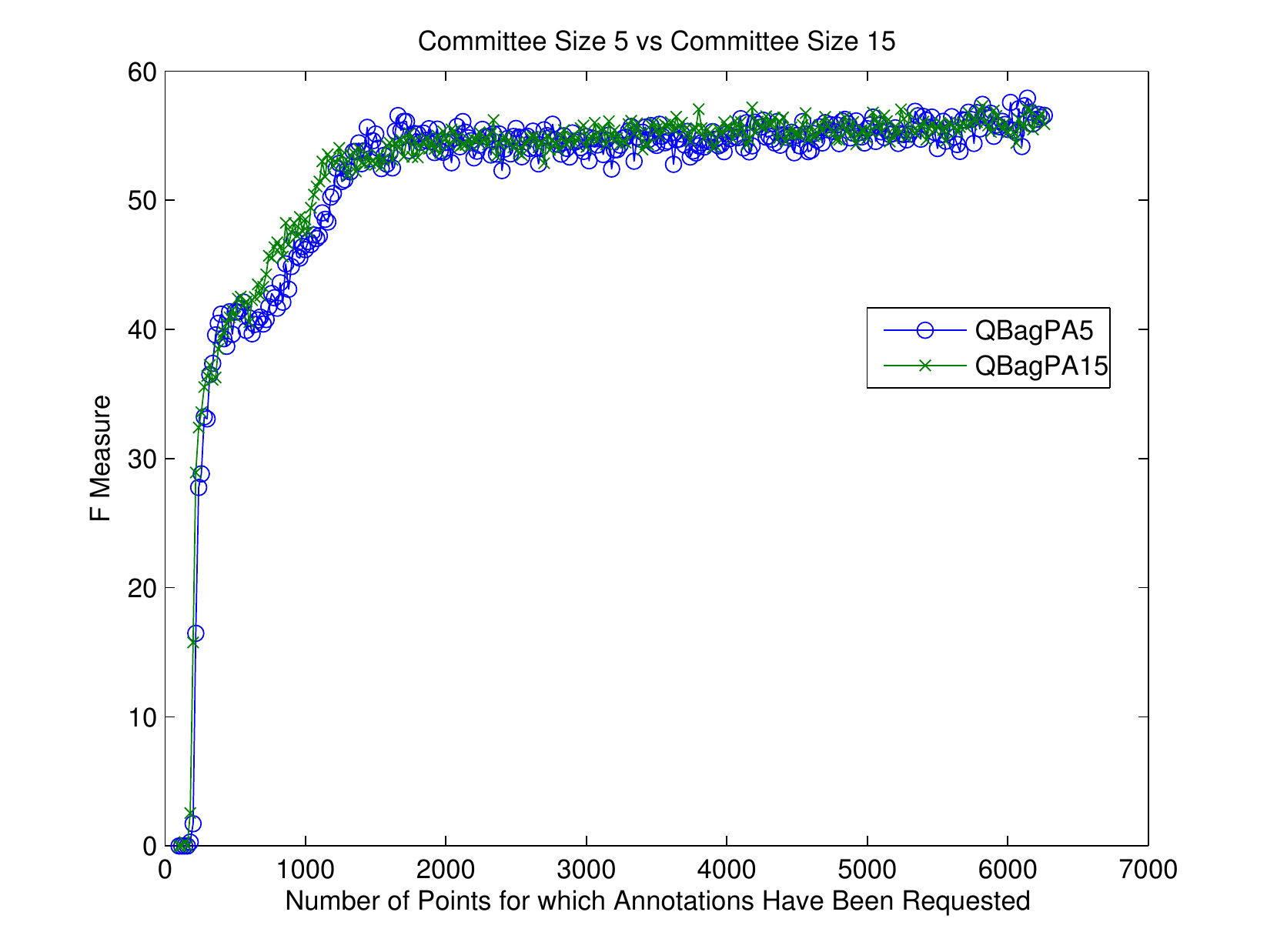}%
\label{fig:BagC5VsBagC15}}
\hfil
\subfloat[QBoostPA with different committee sizes for category Money-Fx from Reuters]{\includegraphics[width=5.7cm]{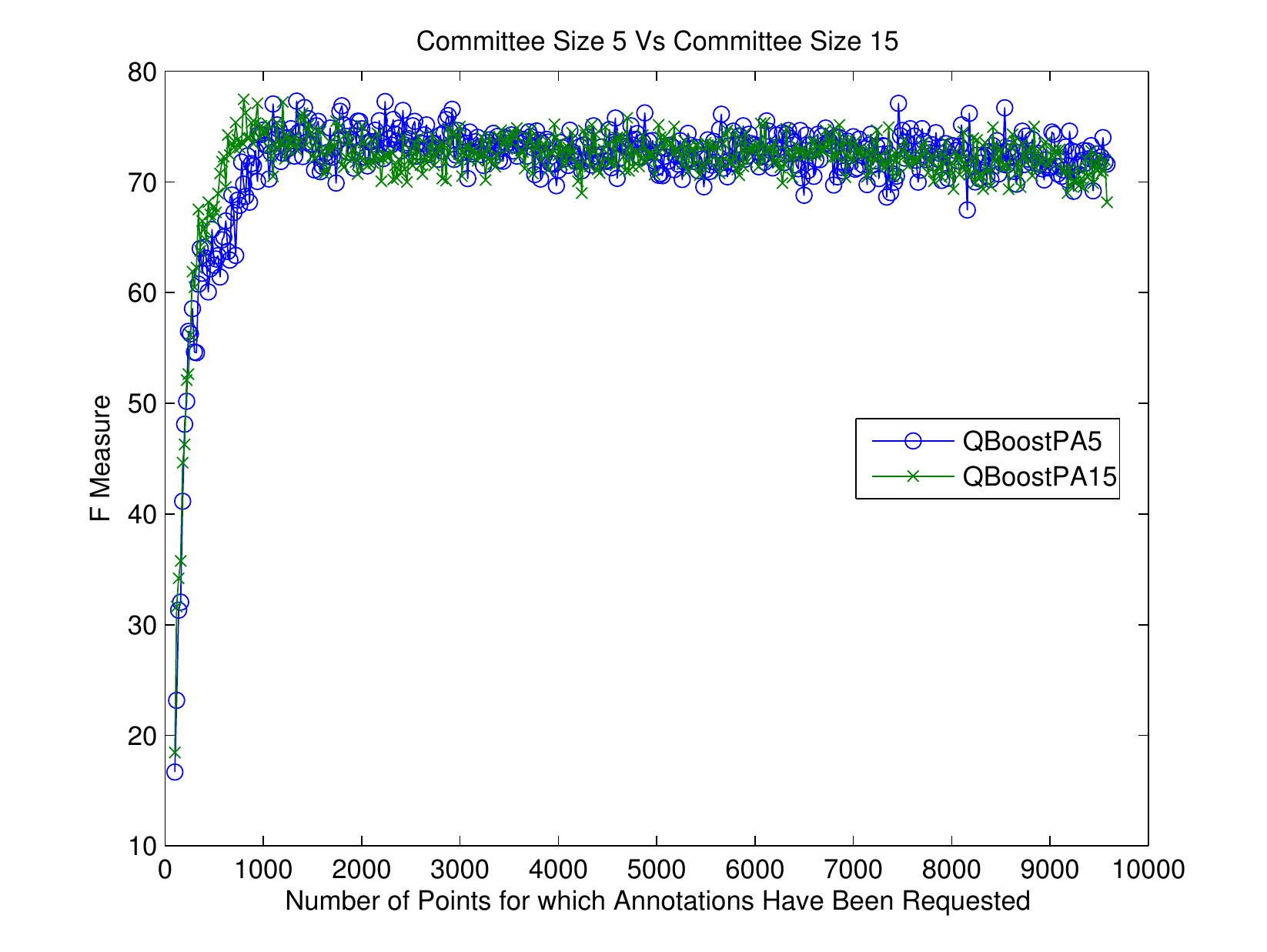}%
\label{fig:BoostC5VsBoostC15}}
\caption{Effects of Committee Size for QBagPA and QBoostPA}
\vspace*{-.2cm}
\label{fig:committeeSizes} 
\end{figure*}

QBC-based approaches have been shown to work slightly better as committee size increases,
with diminishing returns as the committee size is increased \cite{melville2003}.
Figure~\ref{fig:BagC5VsBagC15} shows the performance of QBagPA with a committee of size 5
versus a committee of size 15 for a representative Ohsumed category,
Immunologic Diseases and Fig.~\ref{fig:BoostC5VsBoostC15} shows the performance of QBoostPA
with a committee of size 5 versus a committee of size 15 for a representative
Reuters category, money-fx. Performance on the other categories is similar. Figure \ref{fig:committeeSizes} indicates that increasing the 
committee size from 5 to 15 has only a small improvement for QBagPA and    
QBoostPA. This small gain is not worth the additional computational burden of increasing   
the committee size to 15. 

\section{Discussion} \label{discussion}

This section explores possible reasons why ClosestPA outperforms QBagPA and QBoostPA for 
AL-SVM. Recall
that one of the main goals of AL is to reduce the amount of labeled data by requesting labels for only
useful points, that is, points that will help the base learner learn a better model. During successful
AL, the current set of labeled data $L$ will consist almost exclusively of points that are relevant
for learning and there will be few if any redundant points.\footnote{During the latter stages of an AL
simulation such as the ones we have presented, this may no longer be true as the algorithm is
eventually forced to select redundant examples as the unlabeled pool of data is being exhausted but
these latter stages of AL are irrelevant as AL would have stopped before this in practice.}
We have verified this empirically for our AL-SVM experiments as almost every training point we add
during AL before the stopping area becomes a support vector in the resulting model. 

Consider the following scenario. During earlier rounds of AL a point $x_i$ was selected for annotation
and $(x_i,y_i)$ was placed
into the labeled data $L$. Also, there exists in the current unlabeled pool $U$ a point $x_j$ with
label $y_j = y_i$ that is
very similar to $x_i$. Now a new round of AL is about to occur. 

With ClosestPA, $(x_i,y_i)$ will be used in training and thus it is not likely that $x_j$ will be
selected and instead a different more informative point can be selected. We are not hurt by not
selecting $x_j$ because it's so similar to $x_i$ that the model trained with $(x_i,y_i) \in L$ already
correctly predicts $y_j = y_i$ which is correct. 

With QBagPA/QBoostPA, it's plausible that $(x_i,y_i)$ is sampled for some of the training bags and not
for others. Since $x_i$ was selected during AL, it means the committee disagreed a lot on the label of
$x_i$ when $x_i$ wasn't included in $L$ (the labeled set for training the committee). Since $x_j$ is
similar to $x_i$, the committee members that don't have $x_i$ in their training bag are likely to
disagree a lot on the label of $x_j$. Though the committee members that have $x_i$ in their training
bag are likely to label $x_j$ correctly, overall disagreement on $x_j$ could still be high enough that
$x_j$ gets selected for annotation at the expense of choosing a different point. 

We would now have both
$x_i$ and the similar $x_j$ in $L$, leading to more redundancy and less efficient data utilization than
with ClosestPA to achieve the same level of performance. 

\section{Related Work} \label{relatedWork}

QBC AL was initiated by \cite{seung1992} and made more practically viable by showing how 
the committees could be formed by using boosting/bagging in \cite{abe1998}. Melville and Mooney
\cite{melville2004} use DECORATE \cite{melville2003} to form the committees with decision trees as the
base learners. DECORATE works by generating artificial examples and labeling them according to a
distribution of labels that is inversely proportional to the current ensemble's predictions. DECORATE
is an interesting approach to consider since it helps to create committee diversity. However, the
straightforward way to extend DECORATE for use with SVMs would seem to require explicitly representing
feature vectors for generating the artificial examples. But this goes against one of the important
appealing aspects of using SVMs: the ability to use a very large number of features in a
computationally efficient manner via implicit feature representation via the kernel trick.
Nonetheless, there are many applications of SVMs where the number of features is a manageable finite
number and exploring the performance of QBC AL-SVM with DECORATE for such applications is a 
possibility for future work. McCallum and Nigam \cite{mccallum1998} integrate the Expectation Maximization (EM) algorithm into QBC AL and select queries based on a   
combination of committee disagreement and example density. Using SVMs as the base learner is not
straightforward as their algorithm seems to require a probabilistic method to integrate with their use
of the EM algorithm. None of \cite{seung1992,abe1998,melville2004,mccallum1998} use SVMs as 
the base learner nor do they compare their learners with closest-based active learners. 

AL-SVMs with a closest-based strategy has been explored in
\cite{tong2002,tong2001,campbell2000,schohn2000}. The
only one of these works to compare closest-based AL-SVM with a QBC approach is \cite{tong2002} but 
they only compare with QBC approaches that use Winnow or Naive Bayes as the base learners. They found 
that their closest-based AL-SVM approach worked better than their QBC approaches but one cannot draw 
any conclusions about how the closest-based AL-SVM approach compares with QBC-based {\em SVM} 
approaches. Additionally, none of \cite{tong2002,tong2001,campbell2000,schohn2000} consider class 
imbalance. 

\section{Conclusions} \label{conclusions}

The experimental results show that ClosestPA consistently outperforms the QBoostPA and QBagPA algorithms. ClosestPA is computationally faster than the QBC learners and this is important for AL scenarios so that annotators do not have to wait for the active learner to select examples for which to request labels.
At corresponding points of their learning curves, ClosestPA has higher performance than QBoostPA and QBagPA. Also, at the points around where AL stops, which are the most important ones to examine, ClosestPA outperforms QBagPA and QBoostPA. Lastly, in terms of the data utilization ratio which is often used for evaluating QBC approaches, ClosestPA again achieves superior results.


\section*{Acknowledgment} \label{acknowledgment}

This work was supported in part by The College of New Jersey Support of Scholarly Activities (SOSA) program. 

\bibliographystyle{IEEEtran}
\bibliography{paper}
\end{document}